\documentclass[10pt,twocolumn,letterpaper]{article}

\usepackage{cvpr}              

\newcommand{\subtitle}[1]{{\noindent}{\textbf{#1}}}

\definecolor{cvprblue}{rgb}{0.21,0.49,0.74}
\usepackage[pagebackref,breaklinks,colorlinks,allcolors=cvprblue]{hyperref}

\def\paperID{11341} 
\def\confName{CVPR}
\def\confYear{2026}

\usepackage{booktabs}
\usepackage{multirow}
\usepackage{makecell}
\usepackage{float}
\usepackage[ruled]{algorithm2e} 
\usepackage{algorithmic}

\title{ExpoCM: Exposure-Aware One-Step Generative Single-Image HDR Reconstruction}

\author{
	Aoyu Liu\textsuperscript{\rm1,\footnotemark[1]}, 
    Zhen Liu\textsuperscript{\rm1,\thanks{Equal contribution}}, 
    Ziyi Wang\textsuperscript{\rm1},
    Dian Chen\textsuperscript{\rm1},
    Bing Zeng\textsuperscript{\rm1}, 
    Shuaicheng Liu\textsuperscript{\rm1,\stepcounter{footnote}\thanks{Corresponding author.}} \\
	\textsuperscript{\rm1}University of Electronic Science and Technology of China \\ 
	{\tt\small \{aoyuliu01@std., liuzhen03@std., eezeng@, liushuaicheng@\}uestc.edu.cn}
}

\begin{document}
\maketitle

\begin{abstract}
    Single-image HDR reconstruction aims to recover high dynamic range radiance from a single low dynamic range (LDR) input, but remains highly ill-posed due to detail saturation in over-exposed regions and noise amplification in under-exposed areas. While recent diffusion-based approaches offer powerful generative priors, they often overlook the exposure-dependent nature of the degradation and incur substantial computational costs from iterative sampling. To address these challenges, we propose ExpoCM, a novel one-step generative HDR reconstruction framework that reformulates HDR reconstruction as a Probability Flow ODE (PF-ODE) and constructs exposure-aware consistency trajectories via exposure-dependent perturbations. Specifically, a soft exposure mask is first constructed to separate the LDR image into over-, under-, and well-exposed regions. Based on this partition, region-conditioned consistency trajectories are designed to hallucinate saturated details, suppress noise in dark regions, and preserve reliable structures within a single, distillation-free inference step. To further enhance perceptual quality, we introduce an Exposure-guided Luminance-Chromaticity Loss in the CIE~$\text{L}^*\text{a}^*\text{b}^*$ space, which assigns exposure-aware weights to luminance and chromaticity components, effectively mitigating brightness bias and color drift. Extensive experiments on the HDR-REAL, HDR-EYE, and AIM2025 benchmarks demonstrate that ExpoCM achieves state-of-the-art fidelity and perceptual accuracy, while enabling over 400$\times$ and 20$\times$ faster inference compared to DDPM (1000 steps) and DDIM (50 steps), respectively. Code is available at \url{https://github.com/AoyuLiu01/ExpoCM}.
\end{abstract}

\section{Introduction}
\label{sec:intro}

\begin{figure}[t]
    \centering
    \includegraphics[width=1.0\linewidth]{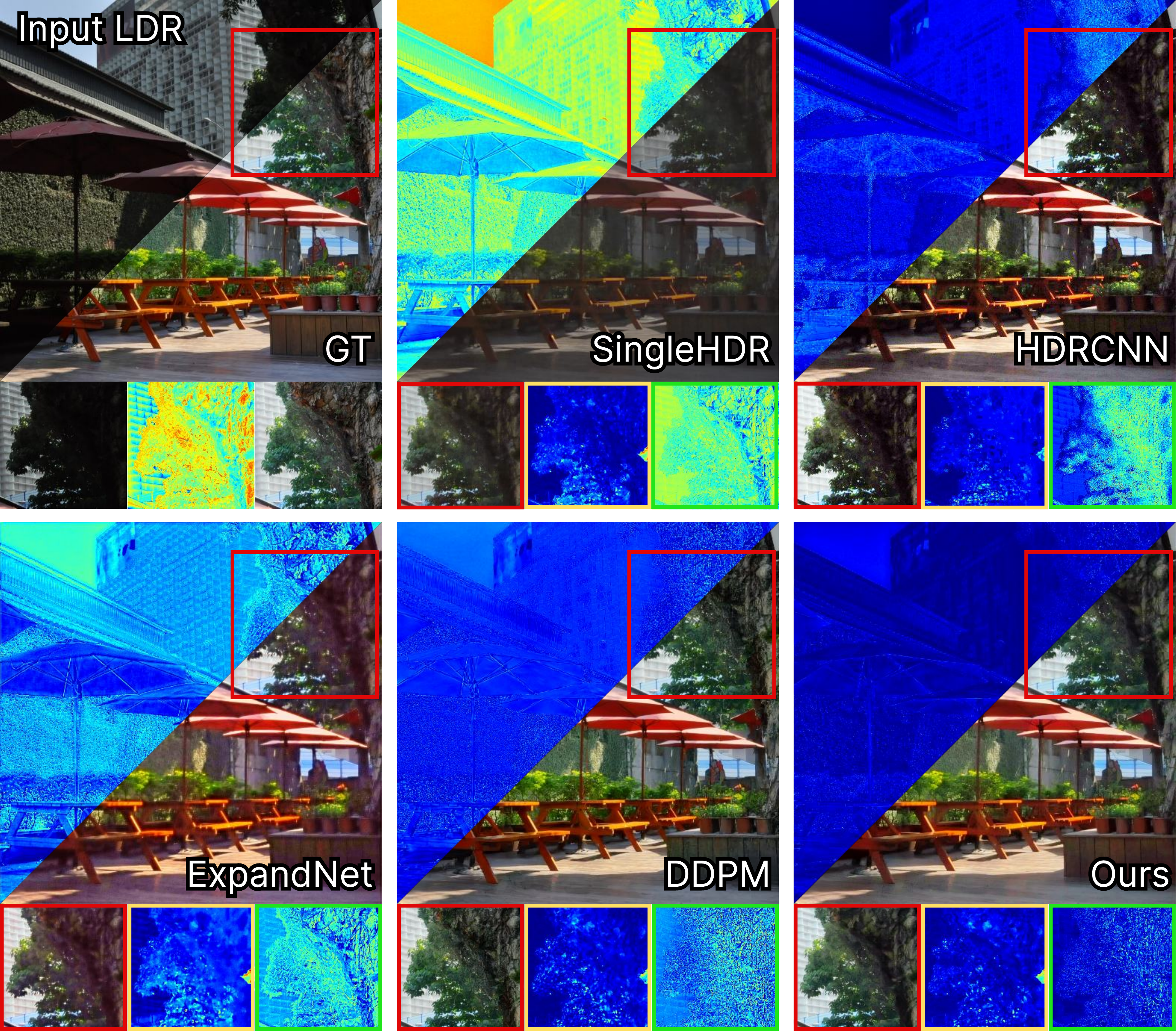}
    \caption{Visual comparisons with previous state-of-the-art methods. For each method, we show the reconstructed result (bottom-right) and the corresponding reconstruction error map (top-left). The highlighted patch (red box) and its associated chrominance error map (yellow box) and luminance error map (green box) are placed below (darker regions indicate smaller errors). The proposed ExpoCM yields results with higher-fidelity global luminance and color information.}
    \label{fig: teaser}
\end{figure}

High dynamic range (HDR) imaging enhances the visual experience of digital content by capturing real-world scene appearances with a wide range of luminance, contrast, and fine details. However, consumer-grade cameras are limited to a narrow dynamic range due to inherent sensor constraints. Substantial research efforts have therefore been devoted to reconstructing HDR images from low dynamic range (LDR) inputs. The most common approach involves fusing multiple LDR images of the same scene at different exposure times~\cite{ExposureFusion,kalantari2017deep,DeepHDR,AHDRNet,ADNet,HDRTransformer}. While effective in static scenes, these methods are susceptible to misalignment and ghosting artifacts in dynamic scenarios due to camera motion or moving objects. To address these issues, another line of research seeks to generate HDR content from a single-exposure image.

Traditional single-image HDR reconstruction methods incorporate hand-crafted priors, such as illumination estimation~\cite{banterle2007framework,banterle2009high} or camera response modeling~\cite{banterle2006inverse,huo2014physiological}, to expand the dynamic range of LDR inputs. In contrast, learning-based solutions~\cite{HDRCNN,HDRUNet,ExpandNet,SingleHDR} primarily reconstruct HDR images in a regression manner using convolutional neural networks (CNNs). However, learning a direct LDR-to-HDR mapping via simple regression is fundamentally ill-posed, as the input LDR image often suffers from severe noise in under-exposed regions and detail saturation in over-exposed areas, leading to suboptimal reconstruction quality.

Recently, generative models, particularly diffusion families, have demonstrated remarkable capability in modeling complex data distributions and achieved impressive success across various low-level vision tasks. However, directly applying diffusion models to single-image HDR reconstruction poses unique challenges. First, the degradation is spatially heterogeneous, as over- and under-exposed areas exhibit distinct visual characteristics. A unified diffusion process struggles to simultaneously recover missing details in saturated regions and suppress noise in dark areas. Second, diffusion models are computationally demanding, typically requiring hundreds of iterative sampling steps to produce high-quality results, leading to high computational costs and limited potential for practical HDR applications.

To address the aforementioned challenges, we propose ExpoCM, an exposure-aware one-step generative framework built upon Consistency Models (CMs), for single-image HDR reconstruction. The core idea is to construct an exposure-aware consistency trajectory that explicitly tailors the generative process to the heterogeneous degradation characteristics within the input LDR image. Specifically, we first introduce an exposure mask generation mechanism to partition the input into well-exposed, under-exposed, and over-exposed regions, each exhibiting distinct information loss patterns. Inspired by recent advances in consistency training, we then formulate HDR reconstruction as a trajectory on the Probability Flow Ordinary Differential Equation (PF-ODE) and further develop an Exposure-Aware Consistency Trajectory, where region-specific perturbations and guidance are injected to tailor the PF-ODE flow according to exposure conditions. This exposure-aware consistency formulation enables high-quality one-step HDR generation without requiring distillation from pre-trained multi-step diffusion models. Moreover, to mitigate the inherent luminance and chromaticity bias of generative models~\cite{diffusion_bias}, we introduce an Exposure-guided Luminance-Chromaticity (ELC) loss in the perceptually uniform CIE L*a*b* space. By assigning exposure-dependent weights to luminance and chromaticity components, the ELC loss adaptively enforces brightness consistency in under-exposed regions and suppresses color drifting in over-exposed areas, leading to more faithful and perceptually accurate HDR reconstruction. To summarize, the main contributions are as follows:
\begin{itemize}
    \item We propose ExpoCM, a novel one-step generative framework for single-image HDR reconstruction that achieves high-fidelity results within a single inference step, eliminating the need for iterative sampling.
    \item We design an Exposure-Aware Consistency Trajectory that tailors the generative process of the Probability Flow ODE (PF-ODE) to the spatially heterogeneous degradations in LDR inputs, which is trained from scratch to avoid the costly distillation process.
    \item We develop an Exposure-guided Luminance-Chromaticity (ELC) loss defined in the perceptually uniform CIE~$\text{L}^*\text{a}^*\text{b}^*$ space. This loss adaptively assigns weights to luminance and chromaticity components based on exposure conditions, effectively mitigating brightness imbalances and color distortion for more perceptually faithful reconstructions.
\end{itemize}

\section{Related Work}
\label{sec:related_work}

\subsection{HDR Reconstruction}
High dynamic range (HDR) image reconstruction can be broadly categorized into multi-image and single-image paradigms according to the number of input LDR images.

\subtitle{Multi-image HDR reconstruction} Early HDR reconstruction methods primarily fuse multiple LDR captures of the same scene, either with bracketed exposures~\cite{ExposureFusion,ma2019deep} or burst sequences~\cite{HDR+}. While this approach produces high-quality results in static scenes, it suffers from misalignment and ghosting artifacts when camera motion or dynamic objects are present. Consequently, a large body of research~\cite{sen2012robust,hu2013hdr,oh2014robust,khan2006ghost} has been devoted to addressing this challenge, commonly referred to as HDR deghosting. Seminal deep learning pipelines~\cite{kalantari2017deep,DeepHDR} established the ``alignment-fusion'' paradigm, where input LDR images are first aligned (e.g., via optical flow~\cite{luo2026learning} or homography~\cite{li2024dmhomo} algorithms) and then fused. Recent efforts extend this by integrating implicit alignment modules~\cite{AHDRNet,ADNet} into end-to-end architectures, or by leveraging advanced designs such as hybrid CNN-ViT networks~\cite{HDRTransformer,chen2023improving,tel2023alignment} and novel learning strategies~\cite{HDR-GAN,prabhakar2021labeled}. Despite these advances, multi-image methods still struggle under severe camera or foreground movements, making them primarily suited for static scenes.

\subtitle{Single-image HDR reconstruction} To overcome the limitations of multi-exposure methods, several studies have explored generating HDR images from a single LDR input. This task is inherently more challenging due to the limited dynamic range and the severe loss of information in over- or under-exposed regions. Early methods relied on hand-crafted priors, such as illumination estimation~\cite{banterle2007framework,banterle2009high} or camera response modeling~\cite{banterle2006inverse,huo2014physiological}. With the advent of deep learning, CNN-based approaches~\cite{HDRCNN,HDRUNet,ExpandNet,SingleHDR,santos2020single,DeepRecursiveHDRI,zhou2020unmodnet} have significantly advanced the field by learning end-to-end LDR-to-HDR mappings. Representative methods like HDRCNN~\cite{HDRCNN} and ExpandNet~\cite{ExpandNet} utilize encoder-decoder architectures, while others like HDRUNet~\cite{HDRUNet} and SingleHDR~\cite{SingleHDR} incorporate imaging priors or multi-branch structures to better recover missing details. Some works~\cite{DeepRecursiveHDRI} first synthesize pseudo multi-exposures from the single input before fusion. Despite their success, these regression-based approaches remain fundamentally limited in addressing the highly ill-posed problem, particularly in recovering details from noise and hallucinating content in saturated areas.

\subsection{Diffusion Models in Low-level Vision}

The remarkable generative capability of diffusion models (DMs)~\cite{DDPM,DDIM} has led to their successful application in various low-level vision tasks, such as image restoration~\cite{DDRM,DiffLL,huang2024detail,li2025exposure,ren2023multiscale,RAW-Flow,cheng2025blind}, inpainting~\cite{repaint,xie2023smartbrush}, and editing~\cite{image_editing1,image_editing2,zhou2024recdiffusion}. However, applying standard DMs to single-image HDR reconstruction presents two major challenges. First, they are computationally expensive, requiring hundreds of iterative sampling steps. Second, the standard diffusion process is spatially-agnostic, treating all pixels uniformly, which is suboptimal for the spatially heterogeneous degradation found in HDR inputs. To address the first challenge, Consistency Models (CMs)~\cite{consistency_model} have recently been proposed to learn a direct mapping from any point on the PF-ODE trajectory to the clean image, enabling high-quality, one-step generation. Nevertheless, these CMs typically inherit the second challenge from their diffusion counterparts. Furthermore, they often introduce a new requirement of being trained via costly distillation from a pre-trained DM. In this work, we bridge these gaps by designing a distillation-free and exposure-aware consistency framework specifically for the HDR reconstruction task.


\begin{figure*}[t]
    \centering
    \includegraphics[width=1.0\linewidth]{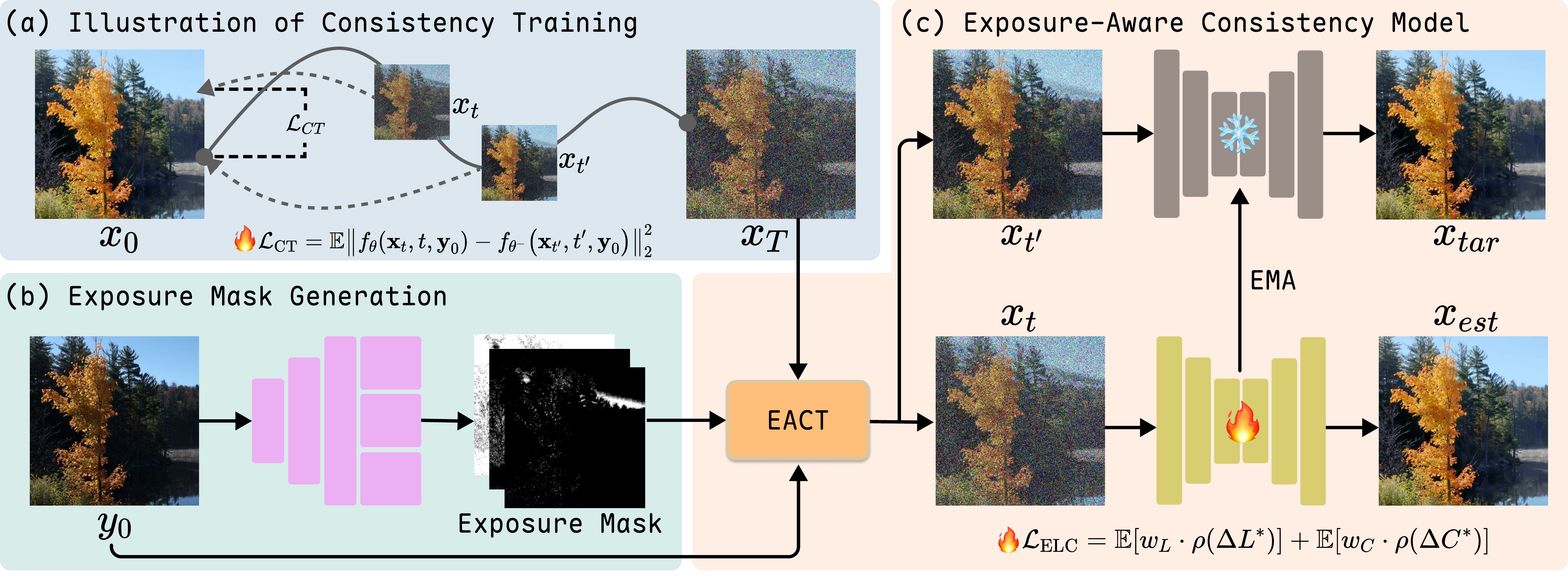}
    \caption{The overall pipeline of our proposed ExpoCM framework. The exposure mask generation module first partitions the input LDR $\mathbf{y}_0$ into over-, under-, and well-exposed regions (Fig.~\ref{fig:pipeline}(b)). Based on these masks, we construct the exposure-aware consistency trajectory (EACT) by formulating and blending three distinct, region-specific generative flows, and the consistency network $f_\theta$ is optimized using the consistency training (CT) loss (Fig.~\ref{fig:pipeline}(a)) and the proposed exposure-guided luminance-chromaticity (ELC) loss (Fig.~\ref{fig:pipeline}(c)).}
    \label{fig:pipeline}
\end{figure*}

\section{Method}
\label{sec:method}

\subsection{Preliminaries}
Given a single LDR image $\mathbf{I}_{L}$, our objective is to reconstruct the corresponding HDR image $\mathbf{I}_{H}$. Unlike previous regression-based approaches, we formulate this task as a conditional generation task built upon Consistency Models (CMs)~\cite{consistency_model} and propose an exposure-aware one-step generative framework, termed ExpoCM, to achieve high-quality HDR reconstruction.

\subtitle{Probability Flow ODE} Let $\mathbf{y}_0$ be the observed LDR image and $\mathbf{x}_0$ be the target HDR image to be reconstructed, diffusion models synthesize data by reversing a forward noising process that perturbs the target HDR image $\mathbf{x}_0$ into a noisy latent state $\mathbf{x}_t$ at time $t \in [0,T]$. This process can be described by a Stochastic Differential Equation (SDE)~\cite{song2020score}:
\begin{equation}
    d\mathbf{x}_t = f(\mathbf{x}_t,t) dt + g(t) d\mathbf{w}_t,
\end{equation}
where $f(\cdot,t)$ and $g(t)$ denote the drift and diffusion coefficients, and $\mathbf{w}_t$ is a standard Wiener process. Correspondingly, there exists a deterministic ordinary differential equation, known as the Probability Flow ODE (PF-ODE), sharing the same marginal probability densities as the SDE:
\begin{equation}
    d\mathbf{x}_t = \Big[f(\mathbf{x}_t,t) - \frac{1}{2}g(t)^2 \nabla_{\mathbf{x}_t} \log p_t(\mathbf{x}_t) \Big]dt.
\end{equation}
Solving this ODE from $t{=}T$ to $t{=}0$ recovers $\mathbf{x}_0$, but numerical integration requires tens or even hundreds of steps.

\subtitle{Conditional Consistency Trajectory} To overcome this limitation, CMs learn a direct mapping from any point $(\mathbf{x}_t, t)$ on the PF-ODE trajectory to its origin $\mathbf{x}_0$. For conditional tasks like HDR reconstruction, the generative trajectory must be guided by the LDR input $\mathbf{y}_0$. A standard approach~\cite{consistency_model} defines the intermediate state $\mathbf{x}_t$ as:
\begin{equation}
\label{eq:baseline_trajectory}
\mathbf{x}_t = (1 - \alpha(t)) \mathbf{x}_0 + \alpha(t) \mathbf{y}_0 + \sigma(t)\boldsymbol{\epsilon},
\end{equation}
where $\alpha(t)$ and $\sigma(t)$ are time-dependent coefficients and $\boldsymbol{\epsilon} \sim \mathcal{N}(0, \mathbf{I})$.

\subtitle{Consistency Training (CT)} As illustrated in Fig.~\ref{fig:pipeline} (a), the CM network $f_\theta$ is trained to predict the trajectory's origin $\mathbf{x}_0$, conditioned on the LDR input $\mathbf{y}_0$. The consistency training objective~\cite{consistency_model} adapted for this conditional task is:
\begin{equation}
    \small 
    \label{eq:baseline_loss}
    \mathcal{L}_{\text{CT}}(\theta,\theta^-) = \mathbb{E}_{\mathbf{x}_0, \mathbf{y}_0, t, t', \boldsymbol{\epsilon}} \left[
\big|\big| f_\theta(\mathbf{x}_{t}, t, \mathbf{y}_0) - f_{\theta^-}(\mathbf{x}_{t'}, t', \mathbf{y}_0) \big|\big|_2^2 \right],
\end{equation}
where $f_\theta$ is the online network, $f_{\theta^-}$ is its exponential moving average (EMA) target, $t' < t$, and $\mathbf{x}_t, \mathbf{x}_{t'}$ are sampled using Eq.~\eqref{eq:baseline_trajectory}. This framework enables one-step inference by computing $\hat{\mathbf{x}}_0 = f_\theta(\mathbf{x}_T, T, \mathbf{y}_0)$, where $\mathbf{x}_T$ can be pure noise or a combination of $\mathbf{y}_0$ and noise.

\subsection{ExpoCM: Exposure-Aware Consistency Framework}
While the baseline conditional trajectory in Eq.~\eqref{eq:baseline_trajectory} enables one-step generation, its formulation implicitly assumes a spatially uniform degradation. It treats all pixels in the LDR input $\mathbf{y}_0$ identically, regardless of their exposure conditions. This uniform treatment hinders its applicability to single-image HDR reconstruction, which is an inherently spatially heterogeneous problem. Specifically, over-exposed regions contain saturated pixels where information is lost and must be hallucinated. Under-exposed regions suffer from severe noise amplification, requiring careful denoising and detail reconstruction. In contrast, well-exposed regions preserve reliable content that should be preserved and enhanced. To address this, we propose an exposure-aware consistency framework, illustrated in Fig.~\ref{fig:pipeline}. The core idea is to replace the single, uniform trajectory (Eq.~\eqref{eq:baseline_trajectory}) with a spatially-varying one that adapts to the local exposure condition. Specifically, as shown in Fig.~\ref{fig:pipeline}(b), we first generate a soft exposure partition of the LDR image through the exposure mask generation module. Then, as depicted in Fig.~\ref{fig:pipeline}(c), we construct a blended exposure-aware consistency trajectory (EACT) tailored to each region's characteristics, which is subsequently used for our exposure-aware consistency training.

\subtitle{Exposure Mask Generation} To distinguish regions with different exposure characteristics in the input LDR image, we construct a soft exposure mask based on luminance statistics. Given an input $\mathbf{I}$, we first compute its luminance channel as $Y = 0.2126 \mathbf{I}_R + 0.7152\mathbf{I}_G + 0.0722 \mathbf{I}_B$. Rather than using fixed thresholds, which are highly sensitive to scene content, we adopt a robust, percentile-based strategy. Specifically, the $2^{\text{nd}}$ and $98^{\text{th}}$ luminance percentiles $(q_{\text{lo}}, q_{\text{hi}})$ are extracted, and a narrow transition band is defined using a margin $\tau{=}0.02$:
\begin{equation}
    l_{\text{core}} = q_{\text{lo}} + \tau (q_{\text{hi}} - q_{\text{lo}}), \quad
    h_{\text{core}} = q_{\text{hi}} - \tau (q_{\text{hi}} - q_{\text{lo}}).
\end{equation}
Pixels darker than $l_{\text{core}}$ are likely under-exposed, while those brighter than $h_{\text{core}}$ are prone to saturation. We thus form continuous low- and high-exposure confidence maps by clipping the normalized distance to this core range:
\begin{equation}
\label{eq:masks_raw}
\begin{aligned}
    m_{\text{low}} &= \text{clip}\left( \frac{l_{\text{core}} - Y}{\tau(q_{\text{hi}} - q_{\text{lo}})}, 0, 1 \right), \\
    m_{\text{high}} &= \text{clip}\left( \frac{Y - h_{\text{core}}}{\tau(q_{\text{hi}} - q_{\text{lo}})}, 0, 1 \right).
\end{aligned}
\end{equation}
Finally, three exposure-aware weighting maps are obtained:
\begin{equation}
\label{eq:masks_final}
    \begin{aligned}
        w_{\text{over}}  &= m_{\text{high}} (1 - m_{\text{low}}), \\
        w_{\text{under}} &= m_{\text{low}} (1 - m_{\text{high}}), \\
        w_{\text{good}}  &= 1 - \max(w_{\text{over}}, w_{\text{under}}).
    \end{aligned}
\end{equation}
These weights softly partition the image into over-exposed, under-exposed, and well-exposed regions. 

\subtitle{Exposure-Aware Consistency Trajectory} After obtaining the exposure masks ${w_{\text{over}}, w_{\text{under}}, w_{\text{good}}}$, we construct region-specific consistency trajectories to align the probability flow with different exposure characteristics. Specifically, in saturated areas, structural information is completely missing in the LDR observation and cannot be reliably recovered from $\mathbf{y}_0$. Therefore, instead of relying on corrupted inputs, we encourage the model to synthesize plausible details purely from noise. The trajectory for over-exposed regions is defined as:
\begin{equation}
\mathbf{x}_t^{o} = (1 - \alpha(t)) \mathbf{x}_0 + \sigma_o(t) \boldsymbol{\epsilon},
\end{equation}
where $\sigma_o(t)$ controls the generation strength. 

In dark areas, the signal is not completely lost but heavily buried in noise. Directly using $\mathbf{y}_0$ introduces amplified noise and blurry details. To provide a reliable yet informative guidance, we apply a low-pass filter to extract coarse structural priors from $\mathbf{y}_0$ and inject them into the trajectory:
\begin{equation}
\mathbf{x}_t^{u} = (1 - \alpha(t)) \mathbf{x}_0 + \alpha(t) \lambda_u \mathcal{F}_{\text{low}}(\mathbf{y}_0) + \sigma_u(t) \boldsymbol{\epsilon},
\end{equation}
where $\mathcal{F}_{\text{low}}(\cdot)$ denotes a low-frequency operator (e.g., Gaussian blur), and $\lambda_u$ adjusts its contribution. For pixels that are neither saturated nor heavily underexposed, the LDR observation remains reliable. Hence, we follow a trajectory similar to the baseline:
\begin{equation}
\mathbf{x}_t^{g} = (1 - \alpha(t)) \mathbf{x}_0 + \alpha(t) \mathbf{y}_0 + \sigma_g(t) \boldsymbol{\epsilon}.
\end{equation}
The full exposure-aware consistency trajectory is obtained by spatially blending the three region-specific trajectories:
\begin{equation}
\label{eq:blended_trajectory}
\mathbf{x}_t =
w_{\text{over}}\odot \mathbf{x}_t^{o} + w_{\text{under}}\odot \mathbf{x}_t^{u} + w_{\text{good}}\odot \mathbf{x}_t^{g},
\end{equation}
where $\odot$ denotes element-wise multiplication. Note that unlike existing exposure-aware generative approaches~\cite{li2025exposure,liu2025solving} that rely on a decoupled two-stage pipeline, which is often slow and artifact-prone, our ExpoCM is a unified one-step framework that solves restoration and generation simultaneously by mathematically blending ODE trajectories.

\subsection{Exposure-guided Luminance-Chromaticity Loss}
While our Exposure-Aware Consistency Trajectory provides a robust generative prior, the framework's perceptual fidelity can be further enhanced. Reconstructed images may still exhibit subtle luminance imbalance or color deviation, especially in severely over- or under-exposed regions. To further enhance perceptual fidelity, we introduce an Exposure-guided Luminance-Chromaticity (ELC) loss. This loss operates in the perceptually uniform CIE~$\text{L}^*\text{a}^*\text{b}^*$ space, which explicitly decouples luminance (L*) from chromaticity (a*, b*), allowing us to apply adaptive, exposure-aware supervision. Given the predicted HDR image $\hat{I}_H$ and ground truth $I_H$, we first convert them into CIE~$\text{L}^*\text{a}^*\text{b}^*$ space:
\begin{equation}
    I_H \rightarrow (L^*, a^*, b^*), \quad 
    \hat I_H \rightarrow (\hat L^*, \hat a^*, \hat b^*).
\end{equation}
The luminance residual is computed as $\Delta L^* = \hat{L}^* - L^*$, and the chromaticity residual as $\Delta C^* = \sqrt{(\hat a^*-a^*)^2 + (\hat b^*-b^*)^2}$.

\subtitle{Exposure-aware weighting strategy.} 
Different exposure regions exhibit distinct reliability in luminance and chromaticity. In under-exposed areas, chromaticity information is unreliable due to noise, whereas luminance still preserves structural cues. Thus, the loss should strongly enforce luminance consistency while reducing the penalty on unreliable chromaticity.  Conversely, in over-exposed regions, pixels saturate towards white, losing color information. Here, the loss must strongly penalize chromaticity errors (i.e., restore color) while being more tolerant of luminance shifts. In well-exposed regions, both components remain reliable and are supervised in a balanced manner.

To implement this behavior in a continuous and differentiable fashion, we design luminance and chromaticity weights $w_L$ and $w_C$ as:
\begin{equation}
    w_L = \lambda_L^{(0)} \left( 
    1 
    + \kappa_L^{\text{lo}} \, s_Y \, w_{\text{under}}^{\alpha}
    + \kappa_L^{\text{hi}} \, A_{\text{spec}} \, w_{\text{over}}^{\alpha}
\right),
\end{equation}
\begin{equation}
    w_C = \lambda_C^{(0)} \left(
    \kappa_C^{\text{hi}} \, w_{\text{over}}^{\alpha} (1 - A_{\text{spec}}) \, h_Y
    + \kappa_C^{\text{lo}} \, w_{\text{under}}^{\alpha} (1 - s_Y)
\right).
\end{equation}
Here, $w_{\text{under}}$ and $w_{\text{over}}$ are the exposure-aware weighting maps defined in Eq.~\eqref{eq:masks_final}. The exponent $\alpha$ controls the sharpness of the mask transitions. $s_Y = \sigma\!\left(\frac{\tau_s - Y}{\delta_s}\right)$ is a shadow-gating function that emphasizes luminance supervision in dark regions. $A_{\text{spec}} = \frac{1}{1 + C_0^*/c_0}$ measures the near-white tendency of highlights (where $C_0^* = \sqrt{a_0^{*2} + b_0^{*2}}$ is the ground-truth chroma and $c_0$ is a scaling constant). $h_Y = \sigma\!\left(\frac{Y - \tau_h}{\delta_h}\right)$ is a highlight visibility factor that modulates chromaticity sensitivity in bright regions. Finally, $\lambda_L^{(0)}$ and $\lambda_C^{(0)}$ are baseline loss weights to ensure stable supervision in well-exposed regions.

Finally, the ELC loss is defined as the weighted sum of robust penalties:
\begin{equation}
    \mathcal{L}_{\text{ELC}} = 
    \mathbb{E} \left[ w_L \cdot \rho(\Delta L^*) \right]
    + \mathbb{E} \left[ w_C \cdot \rho(\Delta C^*) \right],
\end{equation}
where $\rho(\cdot)$ is a robust penalty function (we use the Charbonnier loss, $\rho(x) = \sqrt{x^2 + \epsilon^2}$). We empirically set $\kappa_{L}^{\text{lo}} = 3$, $\kappa_{L}^{\text{hi}} = 1$, $\kappa_{C}^{\text{hi}} = 3$, $\kappa_{C}^{\text{lo}} = 0.5$, $\tau_s = 0.2$, $\tau_h = 0.8$, and $\delta_{\{s,h\}} = 0.1$. Notably, varying $\alpha$, $\kappa_{C}/\kappa_{L}$, or $\tau_s/\tau_h$ slightly results in negligible performance fluctuations ($<0.1$dB).

\subsection{Model Training}
Our consistency network $f_\theta$ is based on the U-Net architecture~\cite{Unet}, utilizing three downsampling and three upsampling stages, with several residual blocks embedded in each. The network's input is formed by concatenating the noisy state $\mathbf{x}_t$ and the LDR image $\mathbf{y}_0$ along the channel dimension. The time step $t$ is converted into a positional embedding and injected into each residual block. We adopt a two-stage training strategy to ensure both generative stability and high perceptual fidelity. In the first stage, we train the network using the consistency training loss ($\mathcal{L}_{\text{CT}}$), as defined in Eq.~\eqref{eq:baseline_loss}, to learn the exposure-aware consistency trajectories and enable stable, one-step inference. In the second stage, we finetune the model using our proposed ELC loss ($\mathcal{L}_{\text{ELC}}$) to explicitly mitigate luminance imbalances and color drift, obtaining the final high-fidelity HDR results. 


\begin{table*}[!t]
    \centering
    \Large
    \caption{Quantitative comparisons on the HDR-REAL~\cite{SingleHDR}, HDR-EYE~\cite{HDR-EYE}, and AIM2025~\cite{AIM} challenge datasets. `-$l$', `-$\mu$', and `-PU' denote metrics computed on linear, $\mu$-law tonemapped, and perceptually-uniform (PU) encoded domain, respectively. $\uparrow$ indicates higher is better, and $\downarrow$ indicates lower is better. The best and second-best results are highlighted in \textbf{bold} and \underline{underlined}, respectively.}
    \resizebox{\linewidth}{!}{
    \begin{tabular}{c|l|cccccccccc}
    \toprule
    \textbf{Dataset} & \textbf{Method} 
    & PSNR-$\mu$ $\uparrow$ & SSIM-$\mu$ $\uparrow$
    & PSNR-PU $\uparrow$ & SSIM-PU $\uparrow$
    & PSNR-$l$ $\uparrow$ & SSIM-$l$ $\uparrow$
    & MS-SSIM $\uparrow$ & HDR-VDP-2/-3 $\uparrow$
    & LPIPS $\downarrow$ & $\Delta E_{2000}$ $\downarrow$ \\
    \midrule
    \multirow{8}[2]{*}{\rotatebox{90}{\textbf{HDR-REAL}~\cite{SingleHDR}}}
        & HDRCNN~\cite{HDRCNN}       & 14.99 & 0.5638 & 16.25 & 0.5305 & 25.81 & 0.5936 & 0.7678 & 31.52\,/\,5.71 & 0.3497 & 21.14 \\
        & SingleHDR~\cite{SingleHDR} & 17.92 & 0.5906 & 19.93 & 0.5695 & 31.08 & 0.7758 & 0.8081 & 34.44\,/\,6.16 & 0.3156 & 14.36 \\
        & ExpandNet~\cite{ExpandNet} & 18.07 & 0.5999 & 20.21 & 0.5953 & 31.70 & 0.7878 & 0.8186 & 30.35\,/\,5.91 & 0.3416 & 13.77 \\
        & HDRUNet~\cite{HDRUNet}     & 16.22 & 0.6327 & 17.79 & 0.6157 & 28.35 & 0.7118 & 0.7293 & 24.23\,/\,4.41 & 0.3937 & 15.72 \\
        & DDPM~\cite{DDPM}           & 25.45 & 0.8173 & 24.78 & 0.7901 & \textbf{37.46} & 0.9063 & 0.9041 & \underline{43.52}\,/\,\underline{7.45} & \underline{0.1921} & 10.40 \\
        & DDIM~\cite{DDIM}           & 20.77 & 0.6925 & 22.65 & 0.6716 & 34.23 & 0.8539 & 0.8307 & 39.14\,/\,6.76 & 0.2365 & 14.26 \\
        & HDR-Trans.~\cite{HDRTransformer} & 15.71 & 0.6027 & 16.83 & 0.5532 & 25.08 & 0.6259 & 0.7870 & 31.72\,/\,5.69 & 0.3665 & 19.24 \\
        & Reti-Diff~\cite{reti-diff} & \underline{27.64} & \underline{0.8354} & \underline{29.15} & \underline{0.8666} & 35.93 & \underline{0.9397} & \underline{0.9147} & 42.08\,/\,7.31 & 0.2645 & \underline{4.83} \\
        & Ours                       & \textbf{28.66} & \textbf{0.8684} & \textbf{30.07} & \textbf{0.8935} & \underline{36.22} & \textbf{0.9521} & \textbf{0.9304} & \textbf{44.27}\,/\,\textbf{7.72} & \textbf{0.1919} & \textbf{4.02} \\
    \midrule
    \multirow{8}[2]{*}{\rotatebox{90}{\textbf{HDR-EYE}~\cite{HDR-EYE}}}
        & HDRCNN~\cite{HDRCNN}       & 15.55 & 0.5986 & 16.12 & 0.5673 & 22.84 & 0.7030 & 0.8049 & 37.84\,/\,7.08 & 0.2811 & 14.05 \\
        & SingleHDR~\cite{SingleHDR} & 15.04 & 0.6535 & 14.36 & 0.5536 & 19.04 & 0.5612 & 0.8813 & 45.23\,/\,7.66 & 0.2436 & 19.28 \\
        & ExpandNet~\cite{ExpandNet} & 16.09 & 0.7023 & 15.15 & 0.6073 & 17.05 & 0.5605 & 0.8878 & 27.97\,/\,7.32 & 0.3105 & 17.55 \\
        & HDRUNet~\cite{HDRUNet}     & 14.81 & 0.6883 & 13.99 & 0.6289 & 17.69 & 0.6014 & 0.8149 & 26.56\,/\,5.79 & 0.3054 & 15.35 \\
        & DDPM~\cite{DDPM}           & 17.45 & \underline{0.7496} & \underline{16.56} & 0.6859 & \textbf{23.38} & 0.6191 & 0.9040 & \underline{53.12}\,/\,\textbf{7.99} & \textbf{0.2005} & 13.81 \\
        & DDIM~\cite{DDIM}           & 16.98 & 0.7647 & 16.03 & \underline{0.7062} & \underline{21.57} & 0.6270 & \underline{0.9044} & \textbf{53.47}\,/\,7.92 & \underline{0.2007} & 13.19 \\
        & HDR-Trans.~\cite{HDRTransformer}& 17.23 & 0.7453 & 16.47 & 0.6889 & 20.85 & 0.6576 & 0.8801 & 44.62\,/\,7.51 & 0.2537 & 12.78 \\
        & Reti-Diff~\cite{reti-diff} & 15.36 & 0.6944 & 14.97 & 0.6163 & 18.77 & 0.5626 & 0.8974 & 46.26\,/\,7.74 & 0.2475 & 17.52 \\
        & Ours                       & \textbf{20.75} & \textbf{0.8017} & \textbf{19.32} & \textbf{0.7638} & 21.30 & \textbf{0.7424} & \textbf{0.9053} & 44.09\,/\,\underline{7.94} & 0.2353 & \textbf{9.68} \\
    \midrule
    \multirow{8}[2]{*}{\rotatebox{90}{\textbf{AIM2025}~\cite{AIM}}}
        & HDRCNN~\cite{HDRCNN}       & 17.67 & 0.6147 & 16.73 & 0.6250 & 23.96 & 0.6750 & 0.8097 & 42.16\,/\,6.57 & 0.3605 & 10.92 \\
        & SingleHDR~\cite{SingleHDR} & 20.77 & 0.7328 & 21.55 & 0.7053 & 28.48 & 0.7930 & 0.9027 & 64.48\,/\,7.90 & 0.2460 & 10.64 \\
        & ExpandNet~\cite{ExpandNet} & 24.94 & 0.8281 & 25.06 & 0.8402 & 29.16 & 0.8765 & 0.9455 & 66.94\,/\,8.21 & 0.2149 & 5.93 \\
        & HDRUNet~\cite{HDRUNet}     & \underline{25.88} & \underline{0.8709} & \underline{26.38} & \underline{0.8808} & \underline{30.98} & \underline{0.9131} & 0.9312 & 57.83\,/\,7.06 & 0.2218 & \underline{4.46} \\
        & DDPM~\cite{DDPM}           & 23.03 & 0.8320 & 23.36 & 0.8145 & 29.71 & 0.8316 & 0.9550 & \textbf{75.57}\,/\,\textbf{8.78} & \textbf{0.1286} & 7.91 \\
        & DDIM~\cite{DDIM}           & 19.50 & 0.7733 & 19.72 & 0.7277 & 27.22 & 0.7602 & 0.9275 & 69.96\,/\,8.36 & 0.1580 & 11.24 \\
        & HDR-Trans.~\cite{HDRTransformer} & 17.12 & 0.7253 & 16.70 & 0.6702 & 20.26 & 0.6174 & 0.8978 & 62.66\,/\,7.78 & 0.2604 & 14.94 \\
        & Reti-Diff~\cite{reti-diff} & 23.31 & 0.8269 & 23.49 & 0.8229 & 28.65 & 0.8370 & 0.9363 & 66.93\,/\,8.19 & 0.2332 & 7.74 \\
        & Ours                       & \textbf{29.02} & \textbf{0.8922} & \textbf{29.06} & \textbf{0.9069} & \textbf{31.97} & \textbf{0.9360} & \textbf{0.9654} & \underline{74.01}\,/\,\underline{8.68} & \underline{0.1511} & \textbf{3.90} \\
    \bottomrule
    \end{tabular}}
    \label{tab:quantitative_results}
\end{table*}

\section{Experiments}
\label{sec:experiments}

\subsection{Experimental Settings}
\label{sec:implementation_details}
\subtitle{Implementation Details} We implement our framework in PyTorch and train our models on NVIDIA 3090 GPUs. The network is trained for 500,000 iterations with a total batch size of 4, using randomly cropped $256 \times 256$ patches. We employ the AdamW optimizer with $\beta_1=0.9$, $\beta_2=0.999$, and $\epsilon=1 \times 10^{-8}$. The initial learning rate is set to $5 \times 10^{-5}$ and is gradually decayed to a minimum of $1 \times 10^{-7}$ using a cosine annealing scheduler. For inference, our ExpoCM processes a $512 \times 512$ image in 0.33s, which is significantly faster than DDPM (174.10s) and DDIM (7.85s).

\subtitle{Datasets} Following~\cite{SingleHDR,dalal2023single}, we conduct our experiments on three widely used benchmarks: HDR-REAL~\cite{SingleHDR}, HDR-EYE~\cite{HDR-EYE}, and the AIM2025 \cite{AIM} Challenge dataset. Specifically, HDR-REAL and HDR-EYE contain 1,838 and 46 LDR-HDR pairs of size 512$\times$512, respectively, while the AIM2025 dataset consists of 18,898 paired samples.

\subtitle{Compared Methods} To assess the performance of the proposed ExpoCM, we compare it with state-of-the-art methods, including HDRCNN~\cite{HDRCNN}, SingleHDR~\cite{SingleHDR}, ExpandNet~\cite{ExpandNet}, HDRUNet~\cite{HDRUNet}, HDR-Transformer~\cite{HDRTransformer}, and Reti-Diff~\cite{reti-diff}. In addition, we also include two representative diffusion models, DDPM~\cite{DDPM} and DDIM~\cite{DDIM}, to demonstrate the advantages of our one-step diffusion framework. 

\subtitle{Evaluation Metrics} We evaluate the reconstruction fidelity using five full-reference metrics, including PSNR, SSIM~\cite{SSIM}, MS-SSIM~\cite{MS-SSIM}, HDR-VDP-2~\cite{HDR-VDP-2}, and HDR-VDP-3~\cite{HDR-VDP-3}. For PSNR and SSIM, evaluations are conducted in three domains: the linear domain, the tonemapped domain obtained by $\mu$-law, and the PU21~\cite{PU21} encoded domain, denoted as `-$l$', `-$\mu$', and `-PU', respectively. For the HDR-VDP-2, we use the standard configuration of 30 pixels per degree (PPD) at a viewing distance of 0.5m. In addition, we report LPIPS~\cite{LPIPS} and $\Delta E_{2000}$ to assess the perceptual quality and color accuracy of the reconstructed results.

\begin{figure*}[!t]
    \centering
    \includegraphics[width=1.0\linewidth]{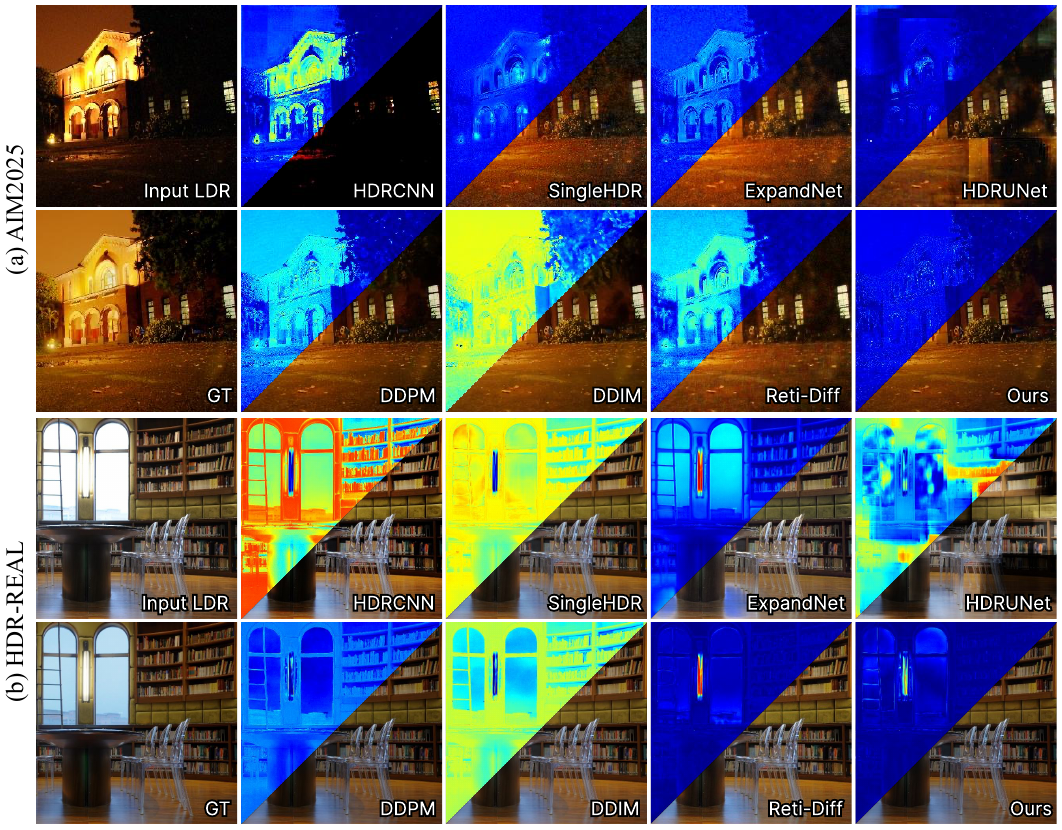}
    \caption{Qualitative comparisons with state-of-the-art single-image HDR reconstruction methods on the AIM2025 and HDR-REAL datasets. For each method, we show the reconstructed HDR image and its corresponding error map, which visualizes the pixel-wise difference from the ground-truth HDR image (darker regions indicate smaller errors).}
    \label{fig:qualitative_results}
\end{figure*}

\subsection{Quantitative Comparison} The quantitative results on the HDR-REAL, HDR-EYE, and the AIM2025 challenge datasets are summarized in Table~\ref{tab:quantitative_results}. Several key observations can be made. First, recent advances in learning-based architectures~\cite{HDRCNN,SingleHDR,ExpandNet,HDRUNet,HDRTransformer,reti-diff} have substantially improved the fidelity of single-image HDR reconstruction, achieving high PSNR and SSIM scores (e.g., HDRUNet~\cite{HDRUNet} and Reti-Diff~\cite{reti-diff}). Second, diffusion-based method DDPM~\cite{DDPM} markedly enhance perceptual quality and obtain the highest LPIPS values, yet often at the expense of pixel-wise fidelity due to their stochastic generation process. Moreover, its efficient variant DDIM~\cite{DDIM} exhibit pronounced performance degradation when the number of sampling steps is reduced, revealing their heavy reliance on iterative denoising. Finally, our proposed ExpoCM, empowered by exposure-aware consistency training, attains state-of-the-art fidelity among single-step methods, while maintaining highly competitive perceptual quality across most datasets with a significantly accelerated inference speed. Furthermore, the incorporation of the proposed Exposure-guided Luminance-Chromaticity (ELC) loss effectively mitigates the inherent luminance and color bias of diffusion models, resulting in more faithful color reconstruction (i.e., the lowest $\Delta E_{2000}$).

\subsection{Qualitative Comparison}

Fig.~\ref{fig:qualitative_results} illustrates the qualitative comparisons with state-of-the-art methods on challenging scenes from the HDR-REAL~\cite{SingleHDR} and AIM2025~\cite{AIM} datasets. As can be seen, previous CNN-based methods (e.g., HDRCNN, SingleHDR, ExpandNet, and HDRUNet) struggle to reconstruct missing details in saturated, over-exposed regions. Furthermore, they tend to produce noticeable blurriness or artifacts in under-exposed areas after denoising. While diffusion-based methods (e.g., DDPM) can alleviate this issue with their strong generative priors, they often suffer from a global brightness bias. Moreover, accelerating these models via reduced sampling steps (e.g., DDIM) significantly degrades reconstruction quality. In contrast, our proposed ExpoCM achieves high-quality HDR reconstruction within a single inference step and can effectively mitigate both global brightness and local color biases, faithfully recovering details in both over- and under-exposed regions.

\subsection{Ablation Study}
In this section, we conduct comprehensive ablation studies to validate the effectiveness of each core component. Unless otherwise specified, all variants are trained following the implementation details in Sec~\ref{sec:implementation_details}. The quantitative results on the HDR-REAL and AIM2025 datasets are reported in Table~\ref{tab: ablation_eact} and Table~\ref{tab: ablation_elc_loss}.

\begin{figure}[t]
    \centering
    \includegraphics[width=1.0\linewidth]{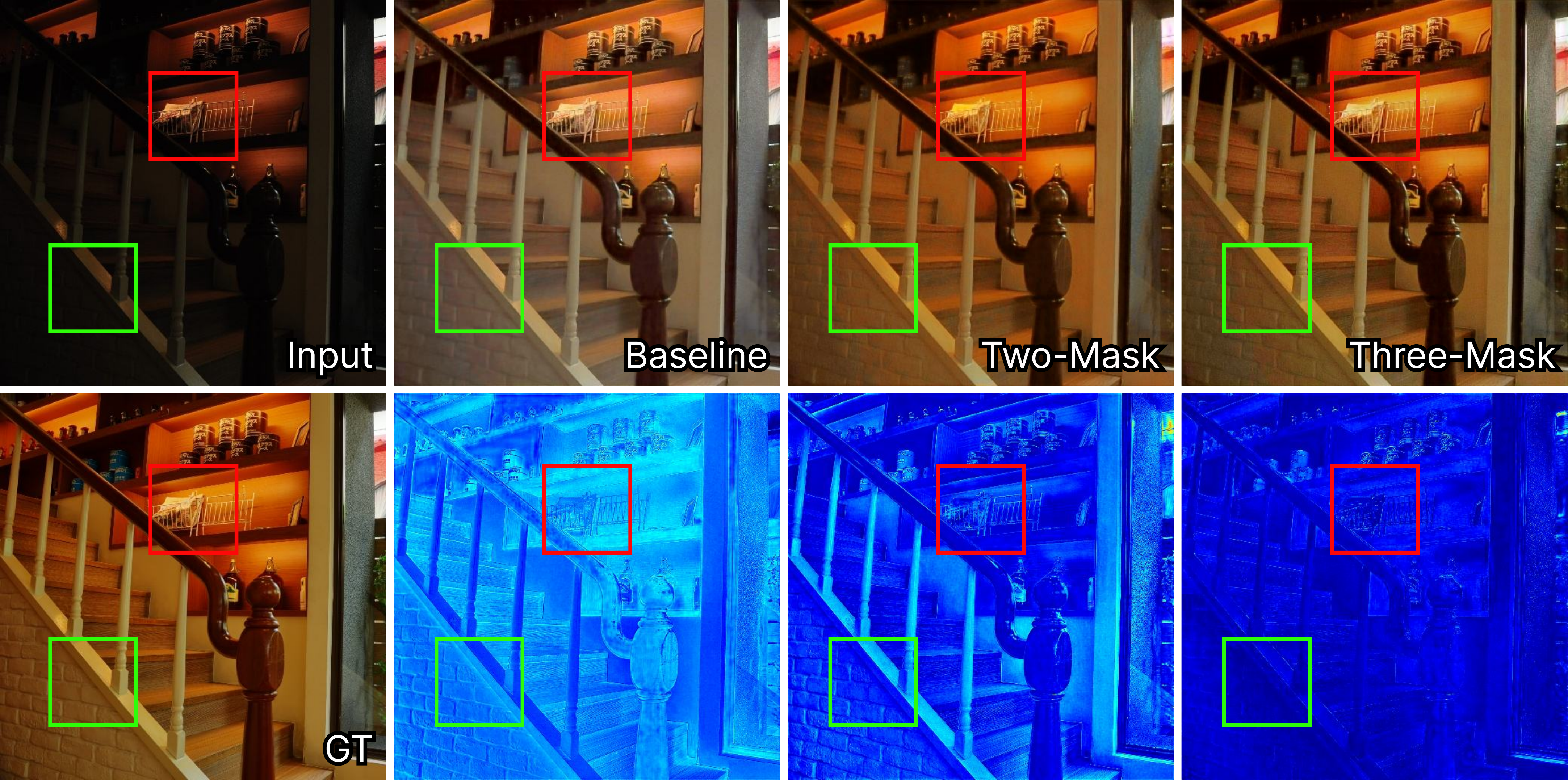}
    \caption{Visual results of our ablation studies on the proposed exposure-aware consistency trajectory. Our full method exhibits the lowest reconstruction error in both the over-exposed (green box) and highlight (red box) regions.}
    \label{fig:ablation_mask}
\end{figure}

\subtitle{Exposure-aware Consistency Trajectory} We first analyze the effectiveness of our core design, the Exposure-Aware Consistency Trajectory. Our central hypothesis is that a spatially heterogeneous trajectory, tailored to regional degradation, is superior to a spatially-agnostic one. To validate this, we compare the three variants detailed in Table~\ref{tab: ablation_eact}: (1) a Baseline model using the uniform trajectory from Eq.~\eqref{eq:baseline_trajectory}, (2) a Two-Mask variant that only distinguishes between well-exposed and ill-posed regions, and (3) our full Three-Mask framework which uses three distinct, region-specific trajectories. As shown in Table~\ref{tab: ablation_eact}, the baseline model yields the poorest results. The Two-Mask variant significantly improves all metrics (e.g., +4.66 PSNR on HDR-REAL), demonstrating the clear benefit of separating reliable from unreliable regions. Our full Three-Mask model achieves the best performance across both datasets. The qualitative results presented in Fig.~\ref{fig:ablation_mask} further reinforce these findings, visually validating our hypothesis that explicitly distinguishing between over-exposure and under-exposure is critical for high-fidelity reconstruction.

\begin{table}[t]
    \centering
    \caption{Quantitative results of the ablation study on our Exposure-Aware Consistency Trajectory (EACT). We compare: (1)  \textbf{Baseline}, which uses a uniform, spatially-agnostic trajectory. (2) \textbf{Two-Mask}, a simplified variant that distinguishes only between well-exposed and ill-posed (over- and under-exposed combined) regions. (3) \textbf{Three-Mask}, our full framework using three distinct trajectories for over-, under-, and well-exposed regions.}
    \Large
    \resizebox{\linewidth}{!}{
      \begin{tabular}{l|ccc|ccc}
      \toprule
      \multirow{2}[4]{*}{Method} & \multicolumn{3}{c|}{HDR-REAL~\cite{SingleHDR}} & \multicolumn{3}{c}{AIM2025~\cite{AIM}} \\
      \cmidrule{2-7} & PSNR-$\mu$ $\uparrow$ & SSIM-$\mu$ $\uparrow$ & LPIPS $\downarrow$ & PSNR-$\mu$ $\uparrow$ & SSIM-$\mu$ $\uparrow$ & LPIPS $\downarrow$ \\
      \midrule     
      Baseline   & 21.09 & 0.6917 & 0.3041 & 27.90 & 0.8842 & 0.1920 \\
      Two-Mask   & 25.75 & 0.8076 & 0.2785 & 28.48 & \textbf{0.8936} & 0.1543 \\
      Three-Mask & \textbf{25.84} & \textbf{0.8282} & \textbf{0.2754} & \textbf{28.89} & 0.8921 & \textbf{0.1504} \\
      \bottomrule
      \end{tabular}}
    \label{tab: ablation_eact}%
\end{table}%

\begin{figure}[t]
    \centering
    \includegraphics[width=1.0\linewidth]{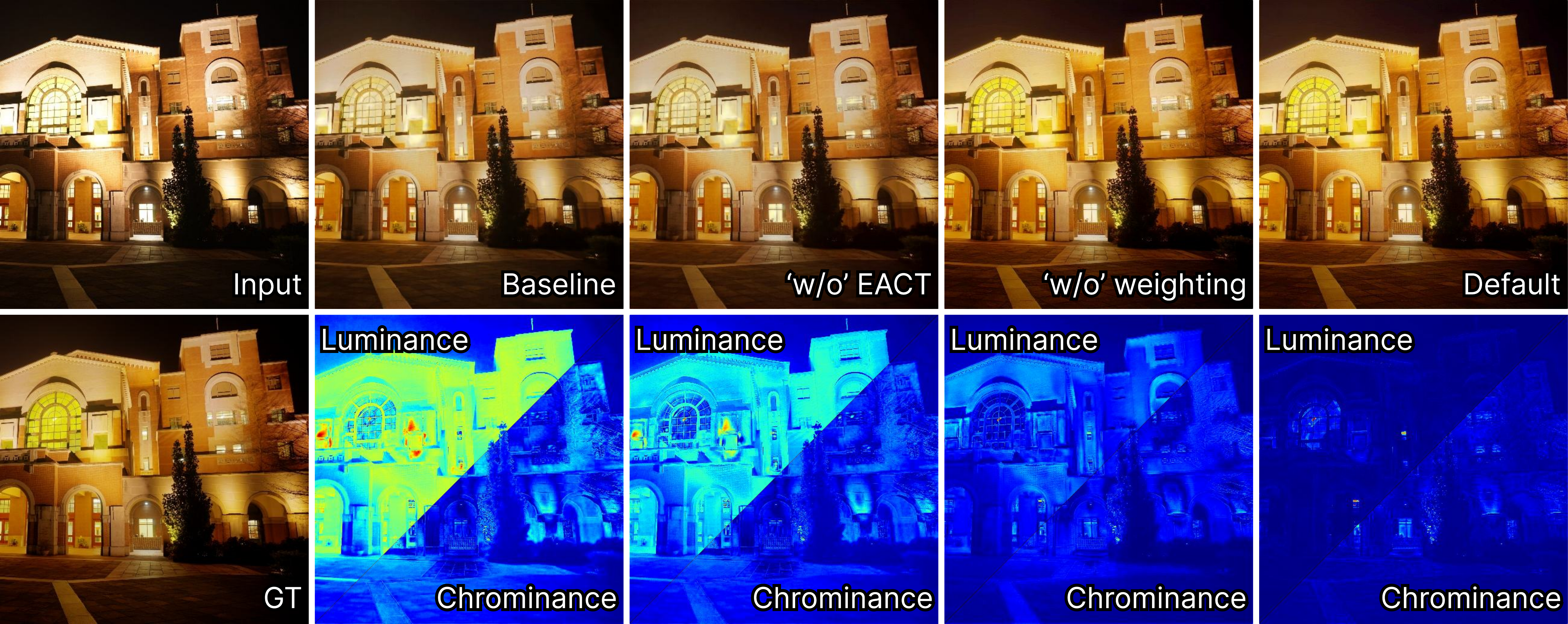}
    \caption{Visual comparisons of our ablation studies on the proposed ELC loss. Compared to all variants, the model optimized with our full ELC loss demonstrates the minimal luminance (top-left) and chrominance (bottom-right) error.}
    \label{fig:ablation_elc}
\end{figure}

\subtitle{Effectiveness of ELC Loss} We next validate the contribution of our Exposure-guided Luminance-Chromaticity (ELC) loss. The quantitative results are presented in Table~\ref{tab: ablation_elc_loss}. We conduct two key comparisons. First, to demonstrate the general effectiveness of the ELC loss, we compare the baseline model with the `w/o EACT' variant. The `w/o EACT' model simply adds our ELC loss onto the baseline trajectory. As shown in the first two rows of Table~\ref{tab: ablation_elc_loss}, this addition provides a consistent improvement across all metrics, most notably reducing the color error $\Delta E_{2000}$ from 12.23 to 12.04 on HDR-REAL. This confirms that the ELC loss enhances perceptual fidelity even on a suboptimal backbone. Second, and more importantly, we isolate the benefit of our exposure-guided weighting strategy. We achieve this by comparing the `w/o weighting' model against our default (i.e., full) framework. Both models utilize our powerful EACT (Three-Mask) trajectory, but the `w/o weighting' variant applies a spatially uniform CIE~$\text{L}^*\text{a}^*\text{b}^*$ loss, (i.e., $w_L$ and $w_C$ are constant). The results clearly show that our default model, with its adaptive weights, outperforms this variant on both datasets, achieving the lowest $\Delta E_{2000}$. The visual comparisons in Fig.~\ref{fig:ablation_elc} further validate the effectiveness of our ELC loss in enforcing decoupled constraints on luminance and chrominance, leading to visibly superior color and brightness fidelity.

\section{Conclusion}
\label{sec:conclusion}

In this paper, we have proposed ExpoCM, a novel one-step generative framework for single-image HDR reconstruction. Our method is designed to address the highly ill-posed nature of this task, specifically the exposure-dependent degradation and the high computational cost of recent generative models. We have introduced an Exposure-Aware Consistency Trajectory (EACT) that partitions the LDR input and tailors the generative PF-ODE flow to over-, under-, and well-exposed regions, enabling high-fidelity reconstruction within a single, distillation-free inference step. To further enhance perceptual quality, we have developed an Exposure-guided Luminance-Chromaticity (ELC) loss in the perceptually uniform CIE~$\text{L}^*\text{a}^*\text{b}^*$ space, which adaptively mitigates brightness imbalance and color drift. Extensive experiments have demonstrated that ExpoCM achieves state-of-the-art fidelity and perceptual accuracy, while being substantially faster than diffusion-based counterparts.

\begin{table}[t]
    \centering
    \caption{Quantitative results of ablation studies about the proposed ELC loss. \textbf{Baseline}: Uniform Trajectory. \textbf{`w/o' EACT}: Uniform Trajectory + Our full ELC loss. \textbf{`w/o' weighting}: EACT Trajectory + Uniform CIE~$\text{L}^*\text{a}^*\text{b}^*$ loss. \textbf{Default} (Ours): EACT Trajectory + Our full ELC loss. Best results are in bold.}
    \Large
    \resizebox{\linewidth}{!}{
      \begin{tabular}{l|ccc|ccc}
      \toprule
      \multirow{2}[4]{*}{Method} & \multicolumn{3}{c|}{HDR-REAL~\cite{SingleHDR}} & \multicolumn{3}{c}{AIM2025~\cite{AIM}} \\
      \cmidrule{2-7} & PSNR-$\mu$ $\uparrow$ & SSIM-$\mu$ $\uparrow$ & $\Delta E_{2000}$ $\downarrow$ & PSNR-$\mu$ $\uparrow$ & SSIM-$\mu$ $\uparrow$ & $\Delta E_{2000}$ $\downarrow$ \\
      \midrule     
      Baseline & 21.09 & 0.6917 & 12.23 & 27.90 & 0.8842 & 4.24 \\
      \midrule
      `w/o' EACT & 21.26 & 0.6988 & 12.04 & 28.23 & 0.8833 & 4.22 \\
      `w/o' weighting & 28.56 & 0.8663 & 4.06 & 28.91 & 0.8898 & 3.95 \\
      \midrule
      Default & \textbf{28.66} & \textbf{0.8684} & \textbf{4.02} & \textbf{29.02} & \textbf{0.8922} & \textbf{3.90} \\
      \bottomrule
      \end{tabular}}
    \label{tab: ablation_elc_loss}%
  \end{table}%

\noindent\textbf{Acknowledgments.} 
This work was supported in part by the National Natural Science Foundation of China (NSFC) under grant 62372091, and in part by the Hainan Province Science and Technology Plan Project under Grant ZDYF2024(LALH)001.

{
    \small
    \bibliographystyle{ieeenat_fullname}
    \bibliography{main}
}


\end{document}